\documentclass[manuscript]{acmart}

\usepackage{multirow}
\usepackage{subfig}
\usepackage{bbold}

\AtBeginDocument{%
  \providecommand\BibTeX{{%
    \normalfont B\kern-0.5em{\scshape i\kern-0.25em b}\kern-0.8em\TeX}}}

\setcopyright{rightsretained}
\begin{document}

\title{Global-Local Item Embedding for Temporal Set Prediction}

\author{Seungjae Jung}
\email{seung.jae.jung@navercorp.com}
\affiliation{%
  \institution{Naver R\&D Center}
  \institution{NAVER CLOVA}
  \city{Seongnam-si}
  \state{Gyeonggi-do}
  \country{Korea, Republic of}
}

\author{Young-Jin Park}
\email{young.j.park@navercorp.com}
\affiliation{
  \institution{Naver R\&D Center}
  \institution{NAVER CLOVA}
  \institution{NAVER AI LAB}
  \city{Seongnam-si}
  \state{Gyeonggi-do}
  \country{Korea, Republic of}
}

\author{Jisu Jeong}
\email{jisu.jeong@navercorp.com}
\author{Kyung-Min Kim}
\email{kyungmin.kim.ml@navercorp.com}
\affiliation{
  \institution{NAVER CLOVA}
  \institution{NAVER AI LAB}
  \city{Seongnam-si}
  \state{Gyeonggi-do}
  \country{Korea, Republic of}
}

\author{Hiun Kim}
\email{hiun.kim@navercorp.com}
\author{Minkyu Kim}
\email{min.kyu.kim@navercorp.com}
\author{Hanock Kwak}
\email{hanock.kwak2@navercorp.com}
\affiliation{%
  \institution{NAVER CLOVA}
  \city{Seongnam-si}
  \state{Gyeonggi-do}
  \country{Korea, Republic of}
}

\renewcommand{\shortauthors}{Jung et al.}
\newcommand{\sj}{\textcolor{blue}}

\begin{abstract}
Temporal set prediction is becoming increasingly important as many companies employ recommender systems in their online businesses, e.g., personalized purchase prediction of shopping baskets.
While most previous techniques have focused on leveraging a user's history, the study of combining it with others' histories remains untapped potential.
This paper proposes \textit{Global-Local Item Embedding} (GLOIE) that learns to utilize the temporal properties of sets across whole users as well as within a user by coining the names as global and local information to distinguish the two temporal patterns.
GLOIE uses Variational Autoencoder (VAE) and dynamic graph-based model to capture global and local information and then applies attention to integrate resulting item embeddings.
Additionally, we propose to use Tweedie output for the decoder of VAE as it can easily model zero-inflated and long-tailed distribution, which is more suitable for several real-world data distributions than Gaussian or multinomial counterparts.
When evaluated on three public benchmarks, our algorithm consistently outperforms previous state-of-the-art methods in most ranking metrics.
\end{abstract}


\copyrightyear{2021}
\acmYear{2021}
\acmConference[RecSys '21]{Fifteenth ACM Conference on Recommender Systems}{September 27-October 1, 2021}{Amsterdam, Netherlands}
\acmBooktitle{Fifteenth ACM Conference on Recommender Systems (RecSys '21), September 27-October 1, 2021, Amsterdam, Netherlands}\acmDOI{10.1145/3460231.3478844}
\acmISBN{978-1-4503-8458-2/21/09}


\maketitle

\section{Introduction}

Many recommendation tasks can be viewed as a problem of predicting the next set given the sequence of sets, e.g., predicting the next basket in online markets and the next playlist in streaming services.
Previous works mainly focused on the local information (a given user's history), applying RNNs \cite{choi2016doctor, yu2016dynamic, hu2019sets2sets} or self-attention \cite{yu2020predicting} to learn the temporal tendency of a sequence of sets within a user.
From the recommender system point of view, temporal set prediction can have a sparsity problem as many users interact with only a small number of items.
Throughout the recommender system literature, the sparsity problem has been dealt with low-rank approximation or collaborative filtering \cite{Sarwar2001Item, Koren2008Factorization, Srebro2005MMMF, Salakhutdinov2007PMF}.
However, such attempts have been less explored in temporal set prediction literature.

In this paper, we propose Global-Local Item Embedding (GLOIE) that integrates global and local information for temporal set prediction.
To capture the global information, we utilize the Variational Autoencoders (VAEs) \cite{Kingma2013VAE, liang2018variational} which are effective on noisy and sparse data.
GLOIE then integrates the local embeddings, that are learned through dynamic graphs \cite{yu2020predicting}, with global embeddings by using an attention method.
In addition, we enhance the performance of GLOIE by using Tweedie distribution for the likelihood of VAE instead of Gaussian or multinomial distributions.
We show that purchase logs follow zero-inflated and long-tail distributions similar to the Tweedie distribution.


We conduct experiments on three public benchmarks, i.e. Dunnhumby Carbo, TaoBao, and Tags-Math-Sx. Empirical results demonstrate that the proposed method outperforms state-of-the-art methods on most metrics.
The main contributions of the paper are summarized as follows:

\begin{figure*}[!tbp]
    \centering
    \includegraphics[width=0.9\textwidth]{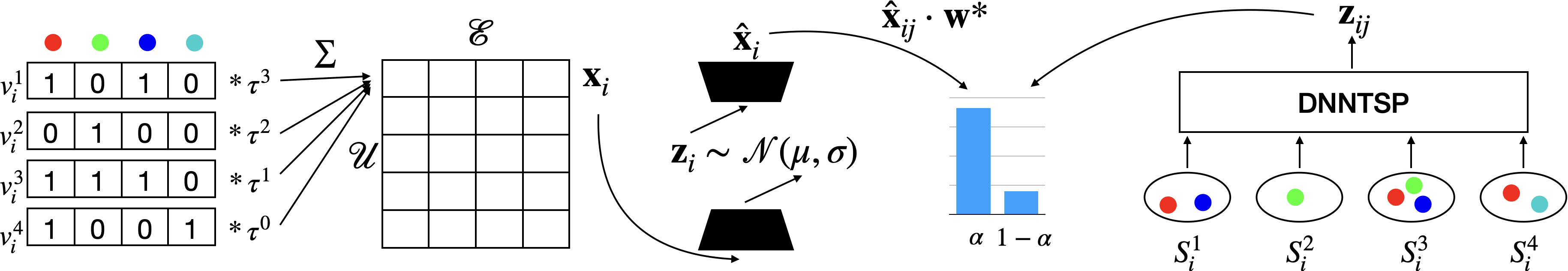}
    \caption{
      Overview of GLOIE.
      We first make sum of time decayed vector $\mathbf{x}_i$ (left part of the figure).
      VAE maximizes the ELBO of each $\mathbf{x}_i$.
      The weighted sum of $\hat{\mathbf{x}}_{ij} \cdot \mathbf{w}^*$ and $\mathbf{z}_{ij}$ becomes final embedding for interacted items.
      The weight $\alpha$ is determined by the attention mechanism.
      For the items that the user never interacted, we just use $\hat{\mathbf{x}}_{ij}$.
    }
    \label{fig:VAE_overview}
    \Description[Overview of GLOIE]{Overview of GLOIE: Sum of time decayed vector, VAE and integration scheme.}
\end{figure*}

\begin{itemize}
    \item We propose GLOIE which differentiate global and local information and integrates them.
    \item We claim that using Tweedie output for VAE decoder is beneficial as it naturally models two properties of data distributions from temporal set prediction problems: zero-inflated and long-tailed.
    \item We achieve state-of-the-art performance on three public benchmarks.
\end{itemize}

\section{Preliminaries}

\subsection{Problem Definition} \label{subsec:problem_definition}

Let  $\mathcal{U} = \{ u_1, u_2, \dots, u_N \}$ be set of users and 
$\mathcal{E} = \{ e_1, e_2, \dots, e_M \}$ be set of items.
Given a user $u_i$'s sequence of sets $\mathbb{S}_i = \{ S_i^k \mid 1 \leq k \leq T_i \}$, our goal is to predict next set $S_i^{T_i + 1}$.
Each set $S_i^k$ can also be represented as a binary vector form $\mathbf{v}_i^k$, where each element $\mathbf{v}_{i, j}^k = \mathbb{1}(e_j \in S_i^k)$.
$\mathbb{1}(x)$ is a indicator function which returns 1 if x is true and 0 otherwise.
We will use the set notation $S_i^k$ and $\mathbf{v}_i^k$ interchangeably to denote user $u_i$'s $k$-th set.

\subsection{Variational Autoencoders}

Variational Autoencoders (VAEs) \citep{Kingma2013VAE, rezende2014stochastic} are a class of deep generative models.
VAEs provide latent structures that can nicely explain the observed data (e.g., customer's purchase history).
Formally speaking, VAEs find the latent variables ($\mathbf{z}$) that maximize the evidence lower bound (ELBO), a surrogate objective function for the maximum likelihood estimation of the given observation $\mathbf{x}$:
\begin{equation} \label{eq:vae_elbo}
    \log p_\theta (\mathbf{x}) \geq
    \mathbb{E}_{q_\phi (\mathbf{z} \mid \mathbf{x})} \left[ \log p_\theta(\mathbf{x}|\mathbf{z}) \right] 
    - D_{KL}(q_\phi(\mathbf{z} \mid \mathbf{x}) || p(\mathbf{z})) 
\end{equation}
where $\theta$ and $\phi$ are model parameters of a decoder and an encoder, respectively.

Encoders are commonly structured by multilayer perceptrons (MLPs) that produce Gaussian distributions over the latent variable $\mathbf{z}$.
On the other hand, decoders are designed to have different probability distributions of output layer depending on the characteristics of datasets.
For example, Gaussian distribution and multinomial distribution are often used to represent the real-valued continuous and binary data, respectively.

\section{Method}

\subsection{Learning Global-Local Information by VAE}

We are interested in modeling other users' history as well as the given user's history.
We proceed with this under the VAE framework.
However, a difficulty arises: every user has a different length of the sequence of sets.
To resolve this, we sum time decayed sequence of sets.

\begin{equation}
\mathbf{x}_i = \sum_{1 \leq k \leq T_i} \mathbf{v}_i^k \cdot \tau^{T_i - k} 
\label{eq:decay}
\end{equation}
where $\mathbf{v}_i^k$ is a vector representation of user $u_i$'s $k$th set $S_i^k$, $\tau \in (0, 1)$ is a decay factor and $T_i$ is the length of the user $u_i$'s sequence as explained in Section \ref{subsec:problem_definition}.
Note that $k$ in $\mathbf{v}_i^k$ represents the index and $T_i-k$ in $\tau^{T_i - k}$ represents the power.
We then update our model to maximize the ELBO of each $x_i$:

\begin{equation} \label{eq:vae_loss}
\mathcal{L}_{vae} =
\mathbb{E}_{\mathbf{x}_i \sim \mathcal{D}_u} \left[
    - \mathbb{E}_{\mathbf{z}_i \sim q(\mathbf{z} \mid \mathbf{x}_i)} [\log p(\mathbf{x}_i \mid \mathbf{z}_i)]
    + D_{KL} (q(\mathbf{z} \mid \mathbf{x}_i) || p(\mathbf{z}))
\right]
\end{equation}
where $\mathcal{D}_u$ is a set of sum of time decayed sequence of sets $\mathbf{x}_i$. 
Equation \eqref{eq:vae_loss} is merely an expectation of negative of Equation \eqref{eq:vae_elbo} over $\mathcal{D}_u$.
Equation \eqref{eq:vae_loss} can be understood as the low-rank approximation with the constraint $D_{KL} (q(\mathbf{z} \mid \mathbf{x}_i) || p(\mathbf{z}))$.

As most people interact with a few items, $\mathbf{v}_i$ contains many $0$s and so does $\mathbf{x}_i$.
However, the reconstructed vector $\hat{\mathbf{x}}_i$ generated from the following process:
\begin{equation} \label{eq:reconstruction}
\mathbf{z}_i \sim q(z \mid \mathbf{x}_i), \quad \hat{\mathbf{x}}_i \sim p(\mathbf{x}_i \mid \mathbf{z}_i) 
\end{equation}
is a dense vector and contains the information of expected preference of user $u_i$ to item $e_j$ in $\hat{\mathbf{x}}_{ij}$ even though the user $u_i$ never interacted with item $e_j$.
Consider two users $u_{i_1}$ and $u_{i_2}$ with similar histories.
$\mathbf{z}_{i_1}$ and $\mathbf{z}_{i_2}$ should be close as $\mathbf{x}_{i_1}$ and $\mathbf{x}_{i_2}$ are close.
In turn, the distance between two reconstructed vectors $\hat{\mathbf{x}}_{i_1}$ and $\hat{\mathbf{x}}_{i_2}$ is small.
Hence, even if user $u_{i_1}$ never interacted with item $e_j$, $\hat{\mathbf{x}}_{i_1 j}$ should be high if $\mathbf{x}_{i_2 j}$ is high.

Note that \citet{hu2020modeling}'s Personalized Item Frequency (PIF) is similar to our sum of time decayed vectors.
However, since their work is based on K-Nearest Neighbor, two shortcomings arise: 1) the inference time grows cubic with the number of users and 2) they cannot use features unlike traditional deep learning approaches.

\subsection{Tweedie Output on Decoder}

\begin{figure*}[!htbp]
    \centering
    \subfloat[$\mu = 0.9, p = 1.3$]{
        \includegraphics[width=0.3\textwidth]{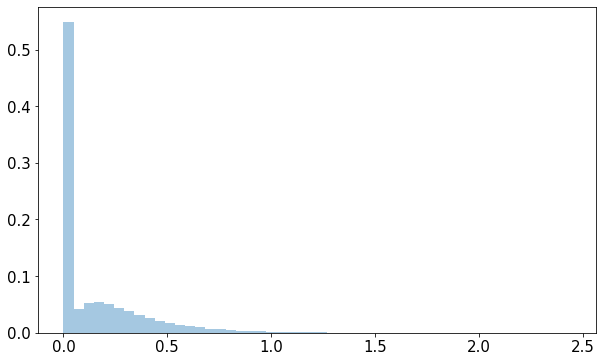}
    }
    \subfloat[$\mu = 1, p = 1.9$]{
        \includegraphics[width=0.3\textwidth]{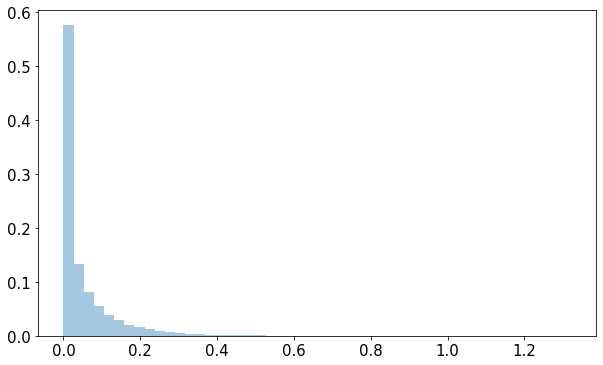}
    }
    \subfloat[$\mu = 0.2, p = 1.9$]{
        \includegraphics[width=0.3\textwidth]{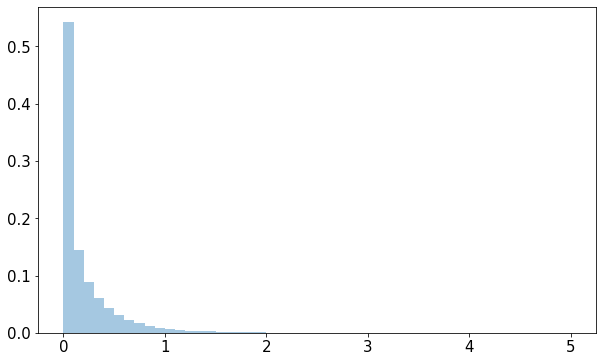}
    }
    \hfill
    \subfloat[DC]{
        \includegraphics[width=0.3\textwidth]{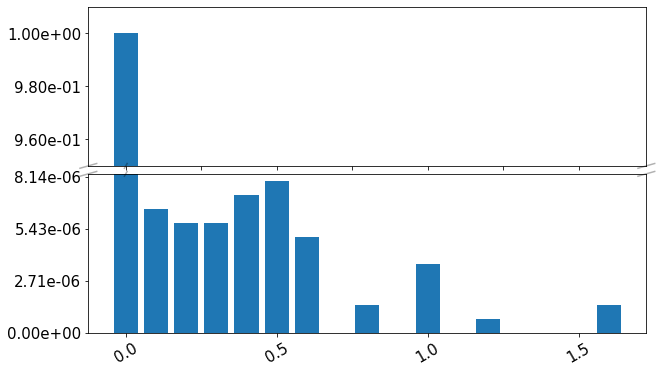}
    }
    \subfloat[TaoBao]{
        \includegraphics[width=0.3\textwidth]{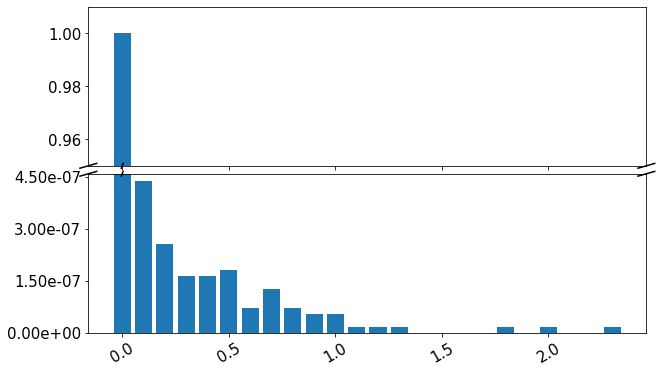}
    }
    \subfloat[TMS]{
        \includegraphics[width=0.3\textwidth]{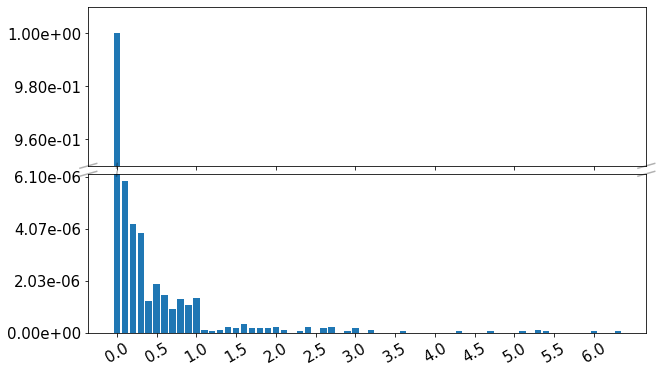}
    }
    \caption{
    Upper row: Histogram plot of samples from Tweedie distribution.
    Lower row: Histogram plot of elements of sum of decayed vector $\mathbf{x}_i$. Across all benchmarks, the distributions are zero-inflated and long-tailed.
    }
    \label{fig:dataset_plots}
\end{figure*}

When training VAE, using gaussian output on $p(\mathbf{x} \mid \mathbf{z})$ is a straightforward option.
However, the distributions of the data generated from temporal set prediction problems are zero-inflated and long-tailed as shown in Figure \ref{fig:dataset_plots}.

Tweedie distribution is a special case of exponential dispersion model (EDM) with a power parameter $p$ and the variance function $V(\mu) = \mu ^ p$ \cite{Jorgensen1987exponential}.
Tweedie distribution with $1 < p < 2$ corresponds to a class of compound Poisson distributions \cite{Jorgensen1997theory}.
Consider two step sampling process $N \sim Poisson(\lambda)$ and $X_i \sim Gamma(\alpha, \beta)$ for $\lambda, \alpha, \beta > 0$ and $i = 1, \dots, N$.
Now we define a random variable $Z$ as follows:

\begin{equation} \label{eq:Tweedie_sample}
Z =
\begin{cases}
0 & N = 0 \\
X_1 + \dots + X_N & N > 0
\end{cases}.
\end{equation}

It is straightforward that the distribution defined by Equation \eqref{eq:Tweedie_sample} is zero-inflated for small enough $\lambda$ and long-tailed as $Z$ is addition of $X_i \sim Gamma(\alpha, \beta)$ for $N > 0$.
Hence using Tweedie output on VAE's decoder for temporal set prediction is beneficial as Tweedie distribution easily captures the properties of distributions that are shown in Figure \ref{fig:dataset_plots}.

Learning the mean parameter $\mu$ and the power parameter $p$ of Tweedie distribution via maximum likelihood is easy.
Minimizing

\begin{equation}
\mathcal{L}_{Tweedie}(z, \mu, p) = - z \cdot \frac{\mu^{1 - p}}{1 - p} + \frac{\mu^{2 - p}}{2 - p}
\end{equation}
maximizes log-likelihood. Here $z$ is the target. See \citet{yang2016insurance} for details.

A line of works chose distributions other than Gaussian or Bernoulli on matrix factorization and VAE \cite{Gopalan2015Scalable, liang2018variational, Truong2021Bilateral}.
Poisson distribution or multinomial distribution are usual choices.
This paper is the first attempt to apply Tweedie distribution for VAE's decoder output to the best of our knowledge.


\subsection{Integrating Global-Local Information}

Though VAE with Tweedie output is already competent, it tends to underestimate a user's preference for the frequently interacted items.
This tendency owes to the learning objective of VAE as the model has to maximize the likelihood of 0 for never interacted items.
This sometimes sacrifices the ability to maximize the likelihood of values of interacted items.
Hence, we integrate item embeddings of frequently interacted items which are learned by state-of-the-art models to our VAE.
We use DNNTSP \cite{yu2020predicting} as it is the state-of-the-art method.
DNNTSP makes user-dependent embeddings of items that are interacted at least once via dynamic graph neural networks.
We denote $\mathbf{z}_{ij}$ as the embedding of user $u_i$ for item $e_j$.

When it comes to combining VAE with embeddings $\mathbf{z}_{ij}$, a problem arises:
the reconstructed value $\hat{\mathbf{x}}_{ij}$ of Equation \eqref{eq:reconstruction} is a scalar while learned embedding $\mathbf{z}_{ij}$ is a vector.
Hence, to match the size between $\hat{\mathbf{x}}_{ij}$ and $\mathbf{z}_{ij}$, we simply multiply a vector $\mathbf{w}^*$, which is of same size as $\mathbf{z}_{ij}$, to $\hat{\mathbf{x}}_{ij}$.
We defer the discussion on the selection of $\mathbf{w}^*$ to latter part of this section.

Now we combine updated embedding with $\hat{\mathbf{x}}_{ij} \cdot \mathbf{w}^*$s and $\mathbf{z}_{ij}$s. The updated embedding $\tilde{\mathbf{z}}_{ij}$ is defined as 

\begin{equation} \label{eq:updated_embedding}
\tilde{\mathbf{z}}_{ij} = 
\begin{cases}
Att(\tilde{\mathbf{x}}_{ij} \cdot \mathbf{w}^*, \mathbf{z}_{ij}) & e_j \in \bigcup_k S_i^k \\
\hat{\mathbf{x}}_{ij} \cdot \mathbf{w}^* & otherwise
\end{cases}
\end{equation}
where $\tilde{\mathbf{x}}_{ij} = \frac{\hat{\mathbf{x}}_{ij}}{\max_j \hat{\mathbf{x}}_{ij}} - 0.5$ which normalizes all values of $\hat{\mathbf{x}}_i$ to $[0.5, -0.5]$ and $Att(\mathbf{q}, \mathbf{k})$ is defined as follows:

\begin{equation}
Att(\mathbf{q}, \mathbf{k}) = \alpha \cdot \mathbf{q} + (1 - \alpha) \cdot \mathbf{k}, 
\, \text{where} \,
\alpha = \sigma \left( (\mathbf{W}_q \cdot \mathbf{q})^T (\mathbf{W}_k \cdot \mathbf{k}) \right).
\label{eq:attn}
\end{equation}

Lastly, we calculate the affinity of a user $u_i$ to item $e_j$ by

\begin{equation} \label{eq:affinity}
\hat{\mathbf{y}}_{ij} = \sigma \left( \tilde{\mathbf{z}}_{ij}^T \mathbf{w}_0 + b_0 \right)
\end{equation}
where $\sigma(x) = \frac{1}{1 + e^{-x}}$.
We choose multi-label soft margin loss for training:

\begin{equation}
\mathcal{L}_{TSP} =
- \frac{1}{N} \sum_i^N \sum_j^M \mathbf{y}_{ij} \log \hat{\mathbf{y}}_{ij} + (1 - \mathbf{y}_{ij}) \log (1 - \hat{\mathbf{y}}_{ij}).
\end{equation}

We now going back to the selection of $\mathbf{w}^*$ in Equation \eqref{eq:updated_embedding}.
We set $\mathbf{w}^* = \mathbf{w}_0$ in Equation \eqref{eq:affinity} though we can set $\mathbf{w}^*$ as a learnable parameter.
If we set $\mathbf{w}^* = \mathbf{w}_0$, Equation \eqref{eq:affinity} becomes $
\hat{\mathbf{y}}_{ij} = \sigma \left( \hat{\mathbf{x}}_{ij} \cdot \mathbf{w}_0^T \mathbf{w}_0 + b_0 \right)
$ for non-interacted item $e_j \notin \bigcup_k S_i^k$, hence it preserves the order of affinity that is learned by VAE.
We also run experiments with learnable parameter $\mathbf{w}^*$ but the performance was on par with $\mathbf{w}^* = \mathbf{w}_0$.

\section{Experiments}

\subsection{Benchmarks}

\begin{table}
  \caption{Statistics of public benchmarks}
  \setlength\tabcolsep{3pt}
  \label{tab:benchmark_statistics}
  \begin{tabular}{ccccccc}
  \toprule
      Dataset & \# of users & \# of sets & \# of elements & \#E/S & \#S/U & \#E/U \\
      \midrule
      DC & 9,010 & 42,905 & 217 & 1.52 & 4.76 & 5.44 \\
      TaoBao & 113,347 & 628,618 & 689 & 1.10 & 5.55 & 4.96 \\
      TMS & 15,726 & 243,394 & 1,565 & 2.19 & 15.48 & 18.05 \\
  \bottomrule
  \end{tabular}
\end{table}

We evaluate our method on three public benchmarks: \textit{Dunnhumby Carbo (DC), TaoBao}, and \textit{Tags-Math-Sx (TMS)}.
We partitioned each dataset into train, validation and test into 70\%, 10\% and 20\% respectively following \citet{yu2020predicting}.
See Table \ref{tab:benchmark_statistics} for statistics of benchmarks.

\subsection{Compared Methods}

We compare four methods: \textit{Toppop, PersonalToppop, Sets2Sets} and \textit{DNNTSP}.

\textit{Toppop} simply serves the items that are interacted the most across all users.
\textit{PersonalToppop} serves the items that the given user interacted with the most.
\textit{Sets2Sets} uses encoder-decoder framework to predict the next set \cite{hu2019sets2sets}. 
Set embeddings are made by pooling operation and set-based attention is used to model temporal correlation relation.
This method also models repeated elements.
\textit{DNNTSP} is composed of three components: Element Relationship Learning (ERL), Temporal Dependency Learning (TDL) and Gated Information Fusing.
ERL is simply a dynamic weighted graph neural networks.
TDL captures temporal dependency.
By Gated Information Fusing each user shares the embeddings of uninteracted items.
Hence DNNTSP can be seen as an ensemble of model which learns local information and \textit{Toppop} model.

\subsection{Results \& Analyses}

The results of our evaluation on three public benchmarks are shown in Table~\ref{tab:experimental_results}.
We consider three metrics: \textit{Recall, Normalized Discounted Cumulative Gain (NDCG)} and \textit{Personal Hit Ratio (PHR)}.
$PHR@K$ is calculated as $\sum_{i=1}^{N^\prime} \mathbb{1} \left( \left| \hat{S}_i \cap S_i \right| > 0 \right) / N^\prime$ where $N^\prime$ is the number of test users, $\hat{S}_i$ is the predicted top-K elements, and $S_i$ is the ground truth set.

\begin{figure*}[!tbp]
    \centering
    \subfloat{
         \includegraphics[width=0.3\textwidth]{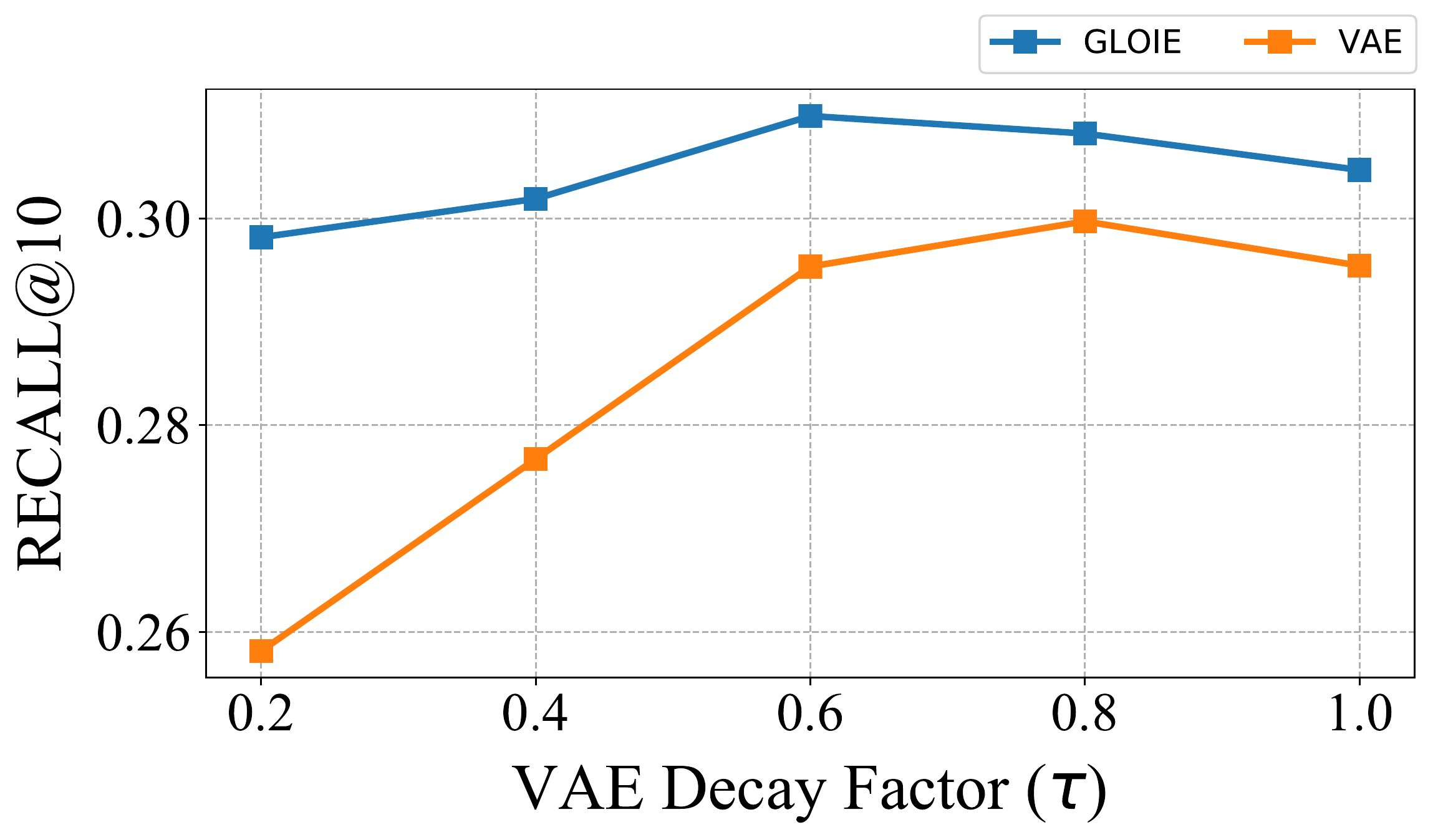}
    }
    \subfloat{
         \includegraphics[width=0.3\textwidth]{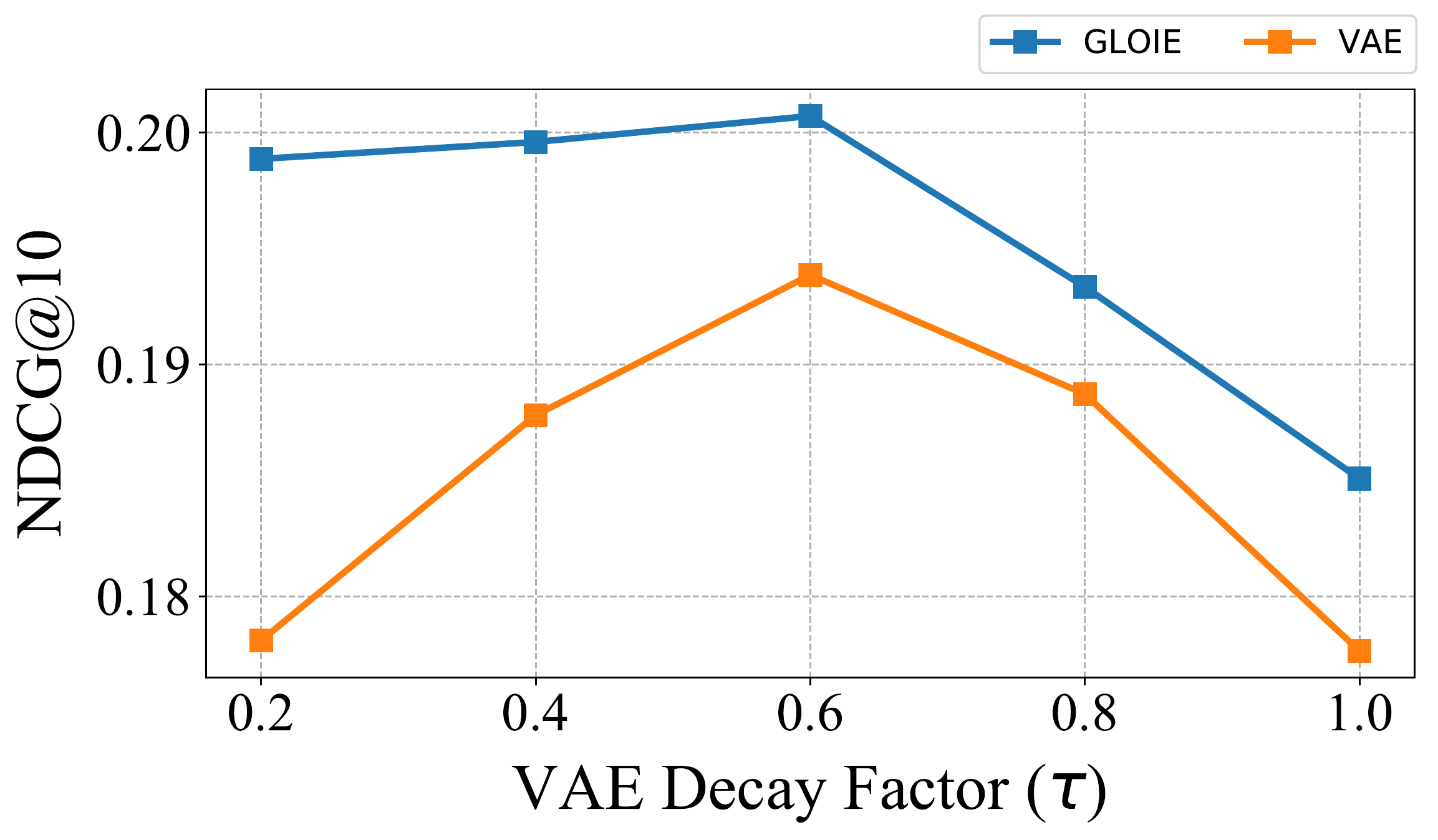}
    }
    \subfloat{
         \includegraphics[width=0.3\textwidth]{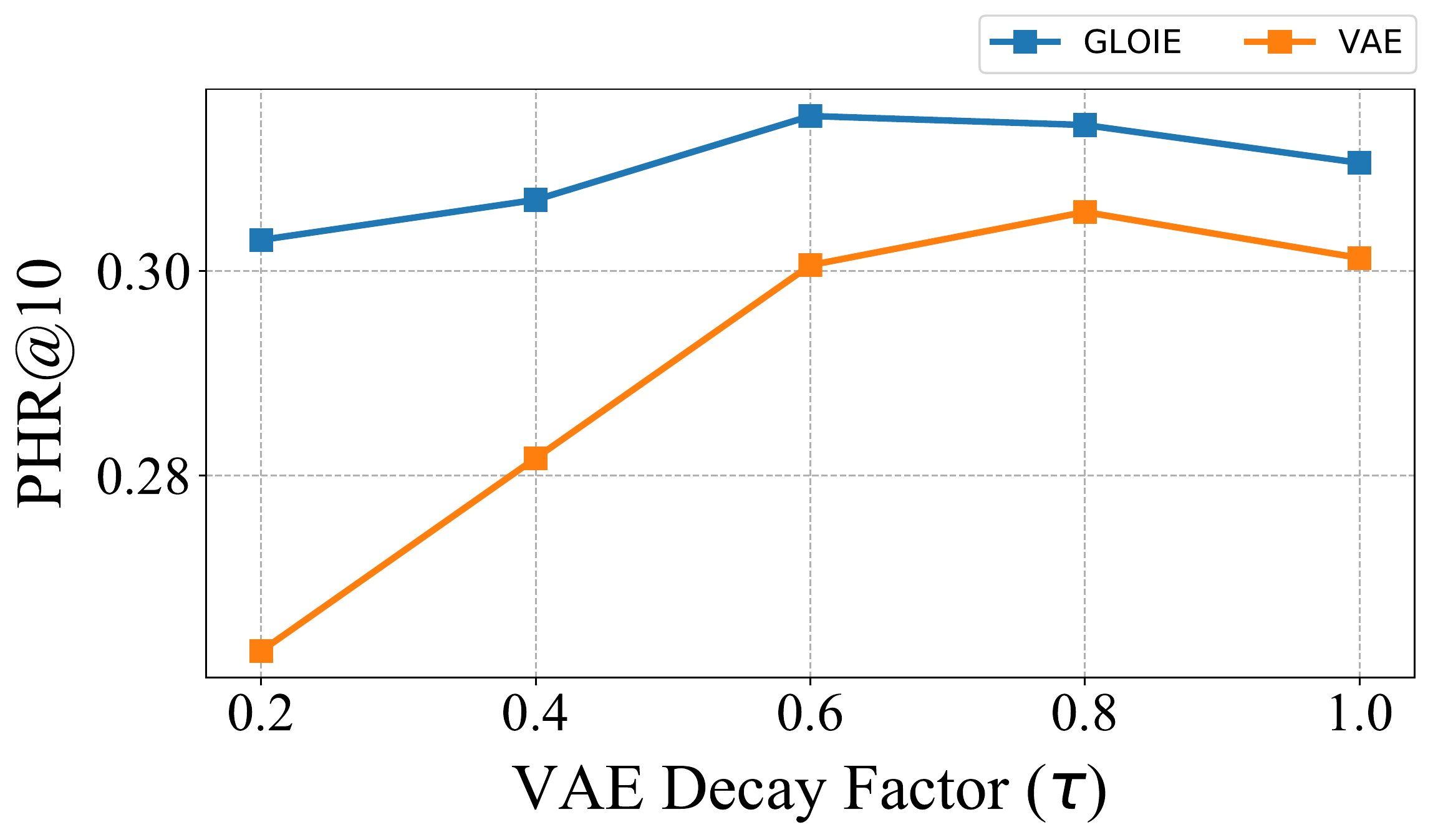}
    }
    \caption{Metric@10 by varying VAE decay factor ($\tau$) on TaoBao dataset.}
    \label{fig:decay_factor}
\end{figure*}

We trained VAE for 30 epochs and then trained DNNTSP for 30 epochs.
We used only one layer for the VAEs across all benchmarks.
For the decay factor in Equation \eqref{eq:decay}, we set $\tau = 0.6$.
As illustrated in Figure \ref{fig:decay_factor}, $\tau=0.6$ shows the best performance on TaoBao dataset.
We empirically observe that $\tau=0.6$ could provide decent results across all metrics on the other datasets as well. 
The dimension of latent space is 128 for DC and TaoBao, and 512 for TMS.
We used Adam optimizer \cite{Kingma2015Adam} with learning rate 0.001.

Across all benchmarks and metrics, GLOIE with Tweedie output outperforms or is on par with all compared methods.
Especially, GLOIE with Tweedie output outperforms the other methods on all metrics on DC and TaoBao datasets.
We can see that GLOIE with Tweedie outperforms DNNTSP on every metric when $K = 10$ which means that the embeddings learned by VAE are richly used as well as the embeddings learned by DNNTSP.

One thing to remark is that VAE with Tweedie output shows comparable performance to all compared methods on most metrics. 
Given that the number of items a user interacted with is 5.44, 4.96, and 18.05 respectively in DC, TaoBao, and TMS, this shows that VAE with Tweedie output captures the preference of users to non-interacted items.

To investigate the effectiveness of the attention-based integration method illustrated in Equation \eqref{eq:attn}, we compared GLOIE with attention to the one with itemwise learnable weight similar to the method proposed in \citet{yu2020predicting}.
For overall datasets and metrics, we could observe performance gains: 3.09\%, 0.45\%, 0.92\% improvement of NDCG@10 on DC, Taobao, and TMS, respectively.

Lastly, we note that the selection of output distribution of decoder on VAE largely affects the performance.
We empirically show that both Gaussian and multinomial outputs do not fit temporal set prediction problems even though Gaussian is a popular choice for VAE and multinomial is a common choice in recommender system literature after the advent of VAECF \cite{liang2018variational}.
\begin{table*}[!t]
    \caption{
    Comparison between various state-of-the-art methods and ours on three public benchmarks.
    All highest scores are in \textbf{bold} and all second best scores are \underline{underlined}.
    }
    \setlength\tabcolsep{4pt}
    \label{tab:experimental_results}
    \begin{tabular}{c|c|ccc|ccc|ccc}
    \toprule
       \multirow{2}{*}{Dataset} & \multirow{2}{*}{Model} & \multicolumn{3}{c|}{k = 10}  & \multicolumn{3}{c|}{k = 20} & \multicolumn{3}{c}{k = 40} \\
        && Recall & NDCG & PHR & Recall & NDCG & PHR & Recall & NDCG & PHR \\
       \midrule
       \multirow{11}{*}{DC} & Toppop & 0.1618 & 0.0880 & 0.2274 & 0.2475 & 0.1116 & 0.3289 & 0.3940 & 0.1448 & 0.4997\\
       & PersonalToppop & 0.4104 & 0.3174 & 0.5031 & 0.4293 & 0.3270 & 0.5258 & 0.4747 & 0.3332 & 0.5785 \\
       & Sets2Sets & 0.4488 & 0.3136 & 0.5458 & 0.5143 & 0.3319 & 0.6162 & 0.6017 & 0.3516 & 0.7005 \\
       & DNNTSP & \underline{0.4564} & \underline{0.3165} & \underline{0.5557} & \underline{0.5294} & \underline{0.3369} & \underline{0.6272} & 0.6180 & \underline{0.3568} & \underline{0.7165} \\
       \cline{2-11}
       & VAE - \textit{Gaussian} & 0.1618 & 0.0882 & 0.2274 & 0.2507 & 0.1128 & 0.3333 & 0.3847 & 0.1430 & 0.4903 \\
       & VAE - \textit{Multinomial} & 0.1602 & 0.0850 & 0.2230 & 0.2492 & 0.1097 & 0.3311 & 0.3767 & 0.1387 & 0.4786 \\
       & VAE - \textit{Tweedie} & 0.4166 & 0.3000 & 0.5108 & 0.5122 & 0.3267 & 0.6062 & \underline{0.6217} & 0.3517 & 0.7088 \\
       & GLOIE - \textit{Gaussian} & 0.3108 & 0.2349 & 0.3971 & 0.3738 & 0.2526 & 0.4664 & 0.4545 & 0.2706 & 0.5563 \\
       & GLOIE - \textit{Multinomial} & 0.3265 & 0.2465 & 0.4143 & 0.3870 & 0.2633 & 0.4798 & 0.4615 & 0.2803 & 0.5602 \\
       & GLOIE - \textit{Tweedie} & \bf 0.4658 & \bf 0.3264 & \bf 0.5613 & \bf 0.5415 & \bf 0.3477 & \bf 0.6351 & \bf 0.6428 & \bf 0.3708 & \bf 0.7288 \\
       \hline
       \hline
       \multirow{11}{*}{TaoBao} & Toppop & 0.1567 & 0.0784 & 0.1613 & 0.2494 & 0.1019 & 0.2545 & 0.3679 & 0.1264 & 0.3745 \\
       & PersonalToppop & 0.2190 & 0.1535 & 0.2230 & 0.2260 & 0.1554 & 0.2306 & 0.2433 & 0.1590 & 0.2484 \\
       & Sets2Sets & 0.2811 & 0.1495 & 0.2868 & 0.3649 & 0.1710 & 0.3713 & 0.4672 & 0.1922 & 0.4739 \\
       & DNNTSP & \underline{0.3035} & 0.1841 & \underline{0.3095} & \underline{0.3811} & 0.2039 & \underline{0.3873} & \underline{0.4776} & 0.2238 & \underline{0.4843} \\
       \cline{2-11}
       & VAE - \textit{Gaussian} & 0.1592 & 0.0750 & 0.1635 & 0.2480 & 0.0974 & 0.2530 & 0.3665 & 0.1219 & 0.3727 \\
       & VAE - \textit{Multinomial} & 0.1588 & 0.0798 & 0.1634 & 0.2494 & 0.1027 & 0.2545 & 0.3660 & 0.1268 & 0.3723 \\
       & VAE - \textit{Tweedie} & 0.2954 & \underline{0.1939} & 0.3006 & 0.3775 & \underline{0.2148} & 0.3827 & 0.4768 & \underline{0.2353} & 0.4822 \\
       & GLOIE - \textit{Gaussian} & 0.2982 & 0.1768 & 0.3044 & 0.3790 & 0.1973 & 0.3851 & 0.4769 & 0.2175 & 0.4835 \\
       & GLOIE - \textit{Multinomial} & 0.2980 & 0.1791 & 0.3040 & 0.3783 & 0.1995 & 0.3846 & 0.4750 & 0.2195 & 0.4819 \\
       & GLOIE - \textit{Tweedie} & \bf 0.3099 & \bf 0.2007 & \bf 0.3152 & \bf 0.3917 & \bf 0.2216 & \bf 0.3972 & \bf 0.4868 & \bf 0.2412 & \bf 0.4924 \\
       \hline
       \hline
       \multirow{11}{*}{TMS} & Toppop & 0.2627 & 0.1627 & 0.4619 & 0.3902 & 0.2017 & 0.6243 & 0.5605 & 0.2448 & 0.8007 \\
       & PersonalToppop & 0.4508 & 0.3464 & 0.6440 & 0.5274 & 0.3721 & 0.7146 & 0.5495 & 0.3771 & 0.7374 \\
       & Sets2Sets & \underline{0.4748} & \underline{0.3782} & \textbf{0.6933} & 0.5601 & \underline{0.4061} & 0.7594 & 0.6627 & \underline{0.4321} & \underline{0.8570} \\
       & DNNTSP & 0.4693 & 0.3473 & 0.6825 & \underline{0.5826} & 0.3839 & \textbf{0.7880} & \underline{0.6840} & 0.4097 & \textbf{0.8748} \\
       \cline{2-11}
       & VAE - \textit{Gaussian} & 0.2731 & 0.1919 & 0.4660 & 0.3913 & 0.2288 & 0.6195 & 0.5496 & 0.2688 & 0.7813 \\
       & VAE - \textit{Multinomial} & 0.2548 & 0.1615 & 0.4431 & 0.3830 & 0.2001 & 0.6020 & 0.5437 & 0.2412 & 0.7740 \\
       & VAE - \textit{Tweedie} & 0.4661 & 0.3744 & 0.6548 & 0.5579 & 0.4040 & 0.7432 & 0.6663 & 0.4316 & 0.8341 \\
       & GLOIE - \textit{Gaussian} & 0.1345 & 0.0833 & 0.2486 & 0.2363 & 0.1155 & 0.4018 & 0.4014 & 0.1570 & 0.6033 \\
       & GLOIE - \textit{Multinomial} & 0.1479 & 0.1029 & 0.2797 & 0.2192 & 0.1252 & 0.3872 & 0.3259 & 0.1524 & 0.5362 \\
       & GLOIE - \textit{Tweedie} & \bf 0.4860 & \bf 0.3823 & \underline{0.6863} & \bf 0.5868 & \bf 0.4144 & \underline{0.7753} & \bf 0.6926 & \bf 0.4418 & 0.8538 \\
       \bottomrule
    \end{tabular}
\end{table*}

\section{Conclusion}

This paper proposes Global-Local Item Embedding (GLOIE) that learns to utilize the temporal properties of sets across whole users as well as within a user.
The proposed model learns global-local information by maximizing ELBO of the sum of time decayed vectors under the VAE framework and integrates local embeddings learned by dynamic graph neural networks.
As users with similar histories are reconstructed to close vectors, we could model the preference of the given user for a non-interacted item if other users with similar histories frequently interacted with the item.
Data analysis and empirical results show that using Tweedie output for VAE's decoder is effective for modeling temporal set prediction.
The proposed method achieves state-of-the-art results by considering global information which is less explored in temporal set prediction literature.

Though the proposed VAE is powerful in itself, there is some room for improvement.
Instead of using the sum of time decayed vector, we can model change of sets in continuous time by Neural ODE \cite{Chen2018NODE, Rubanova2019ODERNN}.
Finding an appropriate form of prior that captures long-tailed distribution for the temporal set prediction can also be a future direction.
Graph modality can also be used \cite{kim2019tripartite}.
Last but not least, searching for better ways to integrating embeddings learned by VAE and embeddings learned by other algorithms which focus on local information is also an important future work.


\bibliographystyle{ACM-Reference-Format}
\bibliography{ref}


\begin{thebibliography}{21}


\ifx \showCODEN    \undefined \def \showCODEN     #1{\unskip}     \fi
\ifx \showDOI      \undefined \def \showDOI       #1{#1}\fi
\ifx \showISBNx    \undefined \def \showISBNx     #1{\unskip}     \fi
\ifx \showISBNxiii \undefined \def \showISBNxiii  #1{\unskip}     \fi
\ifx \showISSN     \undefined \def \showISSN      #1{\unskip}     \fi
\ifx \showLCCN     \undefined \def \showLCCN      #1{\unskip}     \fi
\ifx \shownote     \undefined \def \shownote      #1{#1}          \fi
\ifx \showarticletitle \undefined \def \showarticletitle #1{#1}   \fi
\ifx \showURL      \undefined \def \showURL       {\relax}        \fi
\providecommand\bibfield[2]{#2}
\providecommand\bibinfo[2]{#2}
\providecommand\natexlab[1]{#1}
\providecommand\showeprint[2][]{arXiv:#2}

\bibitem[\protect\citeauthoryear{Chen, Rubanova, Bettencourt, and
  Duvenaud}{Chen et~al\mbox{.}}{2018}]%
        {Chen2018NODE}
\bibfield{author}{\bibinfo{person}{Ricky T.~Q. Chen}, \bibinfo{person}{Yulia
  Rubanova}, \bibinfo{person}{Jesse Bettencourt}, {and}
  \bibinfo{person}{David~K Duvenaud}.} \bibinfo{year}{2018}\natexlab{}.
\newblock \showarticletitle{Neural Ordinary Differential Equations}. In
  \bibinfo{booktitle}{\emph{Advances in Neural Information Processing
  Systems}}, \bibfield{editor}{\bibinfo{person}{S.~Bengio},
  \bibinfo{person}{H.~Wallach}, \bibinfo{person}{H.~Larochelle},
  \bibinfo{person}{K.~Grauman}, \bibinfo{person}{N.~Cesa-Bianchi}, {and}
  \bibinfo{person}{R.~Garnett}} (Eds.), Vol.~\bibinfo{volume}{31}.
  \bibinfo{publisher}{Curran Associates, Inc.}
\newblock
\urldef\tempurl%
\url{https://proceedings.neurips.cc/paper/2018/file/69386f6bb1dfed68692a24c8686939b9-Paper.pdf}
\showURL{%
\tempurl}


\bibitem[\protect\citeauthoryear{Choi, Bahadori, Schuetz, Stewart, and
  Sun}{Choi et~al\mbox{.}}{2016}]%
        {choi2016doctor}
\bibfield{author}{\bibinfo{person}{Edward Choi}, \bibinfo{person}{Mohammad~Taha
  Bahadori}, \bibinfo{person}{Andy Schuetz}, \bibinfo{person}{Walter~F
  Stewart}, {and} \bibinfo{person}{Jimeng Sun}.}
  \bibinfo{year}{2016}\natexlab{}.
\newblock \showarticletitle{Doctor ai: Predicting clinical events via recurrent
  neural networks}. In \bibinfo{booktitle}{\emph{Machine learning for
  healthcare conference}}. PMLR, \bibinfo{pages}{301--318}.
\newblock


\bibitem[\protect\citeauthoryear{Gopalan, Hofman, and Blei}{Gopalan
  et~al\mbox{.}}{2015}]%
        {Gopalan2015Scalable}
\bibfield{author}{\bibinfo{person}{Prem Gopalan}, \bibinfo{person}{Jake~M.
  Hofman}, {and} \bibinfo{person}{David~M. Blei}.}
  \bibinfo{year}{2015}\natexlab{}.
\newblock \showarticletitle{Scalable Recommendation with Hierarchical Poisson
  Factorization}. In \bibinfo{booktitle}{\emph{Proceedings of the Thirty-First
  Conference on Uncertainty in Artificial Intelligence}} (Amsterdam,
  Netherlands) \emph{(\bibinfo{series}{UAI'15})}. \bibinfo{publisher}{AUAI
  Press}, \bibinfo{address}{Arlington, Virginia, USA},
  \bibinfo{pages}{326–335}.
\newblock
\showISBNx{9780996643108}


\bibitem[\protect\citeauthoryear{Hu and He}{Hu and He}{2019}]%
        {hu2019sets2sets}
\bibfield{author}{\bibinfo{person}{Haoji Hu} {and} \bibinfo{person}{Xiangnan
  He}.} \bibinfo{year}{2019}\natexlab{}.
\newblock \showarticletitle{Sets2sets: Learning from sequential sets with
  neural networks}. In \bibinfo{booktitle}{\emph{Proceedings of the 25th ACM
  SIGKDD International Conference on Knowledge Discovery \& Data Mining}}.
  \bibinfo{pages}{1491--1499}.
\newblock


\bibitem[\protect\citeauthoryear{Hu, He, Gao, and Zhang}{Hu
  et~al\mbox{.}}{2020}]%
        {hu2020modeling}
\bibfield{author}{\bibinfo{person}{Haoji Hu}, \bibinfo{person}{Xiangnan He},
  \bibinfo{person}{Jinyang Gao}, {and} \bibinfo{person}{Zhi-Li Zhang}.}
  \bibinfo{year}{2020}\natexlab{}.
\newblock \showarticletitle{Modeling personalized item frequency information
  for next-basket recommendation}. In \bibinfo{booktitle}{\emph{Proceedings of
  the 43rd International ACM SIGIR Conference on Research and Development in
  Information Retrieval}}. \bibinfo{pages}{1071--1080}.
\newblock


\bibitem[\protect\citeauthoryear{J{\o}rgensen}{J{\o}rgensen}{1987}]%
        {Jorgensen1987exponential}
\bibfield{author}{\bibinfo{person}{Bent J{\o}rgensen}.}
  \bibinfo{year}{1987}\natexlab{}.
\newblock \showarticletitle{Exponential Dispersion Models}.
\newblock \bibinfo{journal}{\emph{Journal of the Royal Statistical Society.
  Series B (Methodological)}} \bibinfo{volume}{49}, \bibinfo{number}{2}
  (\bibinfo{year}{1987}), \bibinfo{pages}{127--162}.
\newblock
\showISSN{00359246}
\urldef\tempurl%
\url{http://www.jstor.org/stable/2345415}
\showURL{%
\tempurl}


\bibitem[\protect\citeauthoryear{J{\o}rgensen}{J{\o}rgensen}{1997}]%
        {Jorgensen1997theory}
\bibfield{author}{\bibinfo{person}{Bent J{\o}rgensen}.}
  \bibinfo{year}{1997}\natexlab{}.
\newblock \bibinfo{booktitle}{\emph{The Theory of Dispersion Models}}.
\newblock \bibinfo{publisher}{Chapman and Hall/CRC}.
\newblock


\bibitem[\protect\citeauthoryear{Kim, Kwak, Kwak, Park, Sim, Cho, Kim, Kwon,
  Sung, and Ha}{Kim et~al\mbox{.}}{2019}]%
        {kim2019tripartite}
\bibfield{author}{\bibinfo{person}{Kyung{-}Min Kim},
  \bibinfo{person}{Dong{-}Hyun Kwak}, \bibinfo{person}{Hanock Kwak},
  \bibinfo{person}{Young{-}Jin Park}, \bibinfo{person}{Sangkwon Sim},
  \bibinfo{person}{Jae{-}Han Cho}, \bibinfo{person}{Minkyu Kim},
  \bibinfo{person}{Jihun Kwon}, \bibinfo{person}{Nako Sung}, {and}
  \bibinfo{person}{Jung{-}Woo Ha}.} \bibinfo{year}{2019}\natexlab{}.
\newblock \showarticletitle{Tripartite Heterogeneous Graph Propagation for
  Large-scale Social Recommendation}. In \bibinfo{booktitle}{\emph{Proceedings
  of {ACM} RecSys 2019 Late-Breaking Results}},
  \bibfield{editor}{\bibinfo{person}{Marko Tkalcic} {and} \bibinfo{person}{Sole
  Pera}} (Eds.).
\newblock
\urldef\tempurl%
\url{http://ceur-ws.org/Vol-2431/paper12.pdf}
\showURL{%
\tempurl}


\bibitem[\protect\citeauthoryear{Kingma and Ba}{Kingma and Ba}{2015}]%
        {Kingma2015Adam}
\bibfield{author}{\bibinfo{person}{Diederik~P. Kingma} {and}
  \bibinfo{person}{Jimmy Ba}.} \bibinfo{year}{2015}\natexlab{}.
\newblock \showarticletitle{Adam: {A} Method for Stochastic Optimization}. In
  \bibinfo{booktitle}{\emph{International Conference on Learning
  Representations}}, \bibfield{editor}{\bibinfo{person}{Yoshua Bengio} {and}
  \bibinfo{person}{Yann LeCun}} (Eds.).
\newblock
\urldef\tempurl%
\url{http://arxiv.org/abs/1412.6980}
\showURL{%
\tempurl}


\bibitem[\protect\citeauthoryear{Kingma and Welling}{Kingma and
  Welling}{2014}]%
        {Kingma2013VAE}
\bibfield{author}{\bibinfo{person}{Diederik~P. Kingma} {and}
  \bibinfo{person}{Max Welling}.} \bibinfo{year}{2014}\natexlab{}.
\newblock \showarticletitle{Auto-Encoding Variational Bayes}. In
  \bibinfo{booktitle}{\emph{International Conference on Learning
  Representations}}, \bibfield{editor}{\bibinfo{person}{Yoshua Bengio} {and}
  \bibinfo{person}{Yann LeCun}} (Eds.).
\newblock
\urldef\tempurl%
\url{https://openreview.net/forum?id=33X9fd2-9FyZd}
\showURL{%
\tempurl}


\bibitem[\protect\citeauthoryear{Koren}{Koren}{2008}]%
        {Koren2008Factorization}
\bibfield{author}{\bibinfo{person}{Yehuda Koren}.}
  \bibinfo{year}{2008}\natexlab{}.
\newblock \showarticletitle{Factorization Meets the Neighborhood: A
  Multifaceted Collaborative Filtering Model}. In
  \bibinfo{booktitle}{\emph{Proceedings of the 14th ACM SIGKDD International
  Conference on Knowledge Discovery and Data Mining}} (Las Vegas, Nevada, USA)
  \emph{(\bibinfo{series}{KDD '08})}. \bibinfo{publisher}{Association for
  Computing Machinery}, \bibinfo{address}{New York, NY, USA},
  \bibinfo{pages}{426–434}.
\newblock
\showISBNx{9781605581934}
\urldef\tempurl%
\url{https://doi.org/10.1145/1401890.1401944}
\showDOI{\tempurl}


\bibitem[\protect\citeauthoryear{Liang, Krishnan, Hoffman, and Jebara}{Liang
  et~al\mbox{.}}{2018}]%
        {liang2018variational}
\bibfield{author}{\bibinfo{person}{Dawen Liang}, \bibinfo{person}{Rahul~G
  Krishnan}, \bibinfo{person}{Matthew~D Hoffman}, {and} \bibinfo{person}{Tony
  Jebara}.} \bibinfo{year}{2018}\natexlab{}.
\newblock \showarticletitle{Variational autoencoders for collaborative
  filtering}. In \bibinfo{booktitle}{\emph{Proceedings of the 2018 world wide
  web conference}}. \bibinfo{pages}{689--698}.
\newblock


\bibitem[\protect\citeauthoryear{Rezende, Mohamed, and Wierstra}{Rezende
  et~al\mbox{.}}{2014}]%
        {rezende2014stochastic}
\bibfield{author}{\bibinfo{person}{Danilo~Jimenez Rezende},
  \bibinfo{person}{Shakir Mohamed}, {and} \bibinfo{person}{Daan Wierstra}.}
  \bibinfo{year}{2014}\natexlab{}.
\newblock \showarticletitle{Stochastic backpropagation and approximate
  inference in deep generative models}. In
  \bibinfo{booktitle}{\emph{International conference on machine learning}}.
  PMLR, \bibinfo{pages}{1278--1286}.
\newblock


\bibitem[\protect\citeauthoryear{Rubanova, Chen, and Duvenaud}{Rubanova
  et~al\mbox{.}}{2019}]%
        {Rubanova2019ODERNN}
\bibfield{author}{\bibinfo{person}{Yulia Rubanova}, \bibinfo{person}{Ricky
  T.~Q. Chen}, {and} \bibinfo{person}{David~K Duvenaud}.}
  \bibinfo{year}{2019}\natexlab{}.
\newblock \showarticletitle{Latent Ordinary Differential Equations for
  Irregularly-Sampled Time Series}. In \bibinfo{booktitle}{\emph{Advances in
  Neural Information Processing Systems}},
  \bibfield{editor}{\bibinfo{person}{H.~Wallach},
  \bibinfo{person}{H.~Larochelle}, \bibinfo{person}{A.~Beygelzimer},
  \bibinfo{person}{F.~d\textquotesingle Alch\'{e}-Buc},
  \bibinfo{person}{E.~Fox}, {and} \bibinfo{person}{R.~Garnett}} (Eds.),
  Vol.~\bibinfo{volume}{32}. \bibinfo{publisher}{Curran Associates, Inc.}
\newblock
\urldef\tempurl%
\url{https://proceedings.neurips.cc/paper/2019/file/42a6845a557bef704ad8ac9cb4461d43-Paper.pdf}
\showURL{%
\tempurl}


\bibitem[\protect\citeauthoryear{Salakhutdinov and Mnih}{Salakhutdinov and
  Mnih}{2007}]%
        {Salakhutdinov2007PMF}
\bibfield{author}{\bibinfo{person}{Ruslan Salakhutdinov} {and}
  \bibinfo{person}{Andriy Mnih}.} \bibinfo{year}{2007}\natexlab{}.
\newblock \showarticletitle{Probabilistic Matrix Factorization}. In
  \bibinfo{booktitle}{\emph{Proceedings of the 20th International Conference on
  Neural Information Processing Systems}} (Vancouver, British Columbia, Canada)
  \emph{(\bibinfo{series}{NIPS'07})}. \bibinfo{publisher}{Curran Associates
  Inc.}, \bibinfo{address}{Red Hook, NY, USA}, \bibinfo{pages}{1257–1264}.
\newblock
\showISBNx{9781605603520}


\bibitem[\protect\citeauthoryear{Sarwar, Karypis, Konstan, and Riedl}{Sarwar
  et~al\mbox{.}}{2001}]%
        {Sarwar2001Item}
\bibfield{author}{\bibinfo{person}{Badrul Sarwar}, \bibinfo{person}{George
  Karypis}, \bibinfo{person}{Joseph Konstan}, {and} \bibinfo{person}{John
  Riedl}.} \bibinfo{year}{2001}\natexlab{}.
\newblock \showarticletitle{Item-Based Collaborative Filtering Recommendation
  Algorithms}. In \bibinfo{booktitle}{\emph{Proceedings of the 10th
  International Conference on World Wide Web}} (Hong Kong, Hong Kong)
  \emph{(\bibinfo{series}{WWW '01})}. \bibinfo{publisher}{Association for
  Computing Machinery}, \bibinfo{address}{New York, NY, USA},
  \bibinfo{pages}{285–295}.
\newblock
\showISBNx{1581133480}
\urldef\tempurl%
\url{https://doi.org/10.1145/371920.372071}
\showDOI{\tempurl}


\bibitem[\protect\citeauthoryear{Srebro, Rennie, and Jaakkola}{Srebro
  et~al\mbox{.}}{2005}]%
        {Srebro2005MMMF}
\bibfield{author}{\bibinfo{person}{Nathan Srebro}, \bibinfo{person}{Jason
  Rennie}, {and} \bibinfo{person}{Tommi Jaakkola}.}
  \bibinfo{year}{2005}\natexlab{}.
\newblock \showarticletitle{Maximum-Margin Matrix Factorization}. In
  \bibinfo{booktitle}{\emph{Advances in Neural Information Processing
  Systems}}, \bibfield{editor}{\bibinfo{person}{L.~Saul},
  \bibinfo{person}{Y.~Weiss}, {and} \bibinfo{person}{L.~Bottou}} (Eds.),
  Vol.~\bibinfo{volume}{17}. \bibinfo{publisher}{MIT Press}.
\newblock
\urldef\tempurl%
\url{https://proceedings.neurips.cc/paper/2004/file/e0688d13958a19e087e123148555e4b4-Paper.pdf}
\showURL{%
\tempurl}


\bibitem[\protect\citeauthoryear{Truong, Salah, and Lauw}{Truong
  et~al\mbox{.}}{2021}]%
        {Truong2021Bilateral}
\bibfield{author}{\bibinfo{person}{Quoc-Tuan Truong}, \bibinfo{person}{Aghiles
  Salah}, {and} \bibinfo{person}{Hady~W. Lauw}.}
  \bibinfo{year}{2021}\natexlab{}.
\newblock \showarticletitle{Bilateral Variational Autoencoder for Collaborative
  Filtering}. In \bibinfo{booktitle}{\emph{Proceedings of the 14th ACM
  International Conference on Web Search and Data Mining}} (Virtual Event,
  Israel) \emph{(\bibinfo{series}{WSDM '21})}. \bibinfo{publisher}{Association
  for Computing Machinery}, \bibinfo{address}{New York, NY, USA},
  \bibinfo{pages}{292–300}.
\newblock
\showISBNx{9781450382977}
\urldef\tempurl%
\url{https://doi.org/10.1145/3437963.3441759}
\showDOI{\tempurl}


\bibitem[\protect\citeauthoryear{Yang, Qian, and Zou}{Yang
  et~al\mbox{.}}{2016}]%
        {yang2016insurance}
\bibfield{author}{\bibinfo{person}{Yi Yang}, \bibinfo{person}{Wei Qian}, {and}
  \bibinfo{person}{Hui Zou}.} \bibinfo{year}{2016}\natexlab{}.
\newblock \bibinfo{title}{Insurance Premium Prediction via Gradient
  Tree-Boosted Tweedie Compound Poisson Models}.
\newblock
\newblock
\showeprint[arxiv]{1508.06378}~[stat.ME]


\bibitem[\protect\citeauthoryear{Yu, Liu, Wu, Wang, and Tan}{Yu
  et~al\mbox{.}}{2016}]%
        {yu2016dynamic}
\bibfield{author}{\bibinfo{person}{Feng Yu}, \bibinfo{person}{Qiang Liu},
  \bibinfo{person}{Shu Wu}, \bibinfo{person}{Liang Wang}, {and}
  \bibinfo{person}{Tieniu Tan}.} \bibinfo{year}{2016}\natexlab{}.
\newblock \showarticletitle{A dynamic recurrent model for next basket
  recommendation}. In \bibinfo{booktitle}{\emph{Proceedings of the 39th
  International ACM SIGIR conference on Research and Development in Information
  Retrieval}}. \bibinfo{pages}{729--732}.
\newblock


\bibitem[\protect\citeauthoryear{Yu, Sun, Du, Liu, Xiong, and Lv}{Yu
  et~al\mbox{.}}{2020}]%
        {yu2020predicting}
\bibfield{author}{\bibinfo{person}{Le Yu}, \bibinfo{person}{Leilei Sun},
  \bibinfo{person}{Bowen Du}, \bibinfo{person}{Chuanren Liu},
  \bibinfo{person}{Hui Xiong}, {and} \bibinfo{person}{Weifeng Lv}.}
  \bibinfo{year}{2020}\natexlab{}.
\newblock \showarticletitle{Predicting Temporal Sets with Deep Neural
  Networks}. In \bibinfo{booktitle}{\emph{Proceedings of the 26th ACM SIGKDD
  International Conference on Knowledge Discovery \& Data Mining}}.
  \bibinfo{pages}{1083--1091}.
\newblock


\end{thebibliography}

\appendix

\end{document}